\documentclass{Interspeech2024}
\usepackage{stfloats}

\interspeechcameraready

\title{Exploring the anatomy of articulation rate in spontaneous English speech: relationships between utterance length effects and social factors}

\name[affiliation={1}]{James}{Tanner}
\name[affiliation={2}]{Morgan}{Sonderegger}
\name[affiliation={1}]{Jane}{Stuart-Smith}
\name[affiliation={3}]{Tyler}{Kendall}
\name[affiliation={4}]{Jeff}{Mielke}
\name[affiliation={4}]{Robin}{Dodsworth}
\name[affiliation={4}]{Erik}{Thomas}

\address{
  $^1$University of Glasgow, United Kingdom
  $^2$McGill University, Canada\\
  $^3$University of Oregon, United States
  $^4$North Carolina State University, United States}
\email{\{james.tanner,jane.stuart-smith\}@glasgow.ac.uk, morgan.sonderegger@mcgill.ca, tsk@uoregon.edu, \{jimielke,rmdodswo,erthomas\}@ncsu.edu}


\begin{document}

\maketitle
\begin{abstract}
Speech rate has been shown to vary across social categories such as gender, age, and dialect, while also being conditioned by properties of speech planning. The effect of utterance length, where speech rate is faster and less variable for longer utterances, has also been shown to reduce the role of social factors once it has been accounted for, leaving unclear the relationship between social factors and speech production in conditioning speech rate. Through modelling of speech rate across 13 English speech corpora, it is found that utterance length has the largest effect on speech rate, though this effect itself varies little across corpora and speakers. While age and gender also modulate speech rate, their effects are much smaller in magnitude. These findings suggest utterance length effects may be conditioned by articulatory and perceptual constraints, and that social influences on speech rate should be interpreted in the broader context of how speech rate variation is structured.



\end{abstract}

\section{Introduction}
Speech rate -- the most-studied component of speech timing -- has been found to be affected by both low-level properties of speech production planning as well as social and speaker-specific attributes, such as speaker dialect, age, gender, and style. While these social factors have been the primary focus of the majority of previous work concerning variation in speech rate (perhaps in part due to strong social stereotypes regarding differences in rate \cite{niedzielski2000,harnsberger2008}, such as older speakers having slower rates), studies across a range of languages have observed mixed results regarding the size and robustness of these social effects, with numerous studies observing rate differences between dialects \cite{verhoeven2004,jacewicz2010rate,kendall2013,clopper2015rate,coats2020,cichoki2023}, gender \cite{Jacewicz2009rate,lee2017irish}, age \cite{fougeron2021}, and style \cite{koreman2006,bona2014}, and others observing either small or negligible differences between groups \cite{ray1990,borsel2008,lee2017}. In contrast, the effect of \emph{utterance length} -- where longer utterances result in both higher average speech rate and lower variance in rate -- has been shown to be the strongest predictor of speech rate differences \cite{quene2008,kendall2013,bishop2018}, and is a particular instantiation of the more general `Menzerath's Law' \cite{menzerath1954}, which predicts that the constituent units of a structure (here syllables) are shortened when the structure itself (here utterances) is longer \cite{altmann1980}. In two studies comparing two dialects of Dutch and US English respectively, the size of these social effects were also shown to be reduced when utterance length was accounted for \cite{quene2008,jacewicz2010rate}, and that longer utterances increase speech rate as well as reduce the \emph{variance} in speech rate (i.e. longer utterances have less-variable speech rates). In addition, evidence demonstrating that individual speakers vary substantially in both speech rate and utterance length \cite{tsao1997,tsao2006,quene2008,jacewicz2010rate} indicates that the relationship between the speech production and social factors conditioning speech rate remain unclear, particularly with respect to how utterance length effects may structurally differ across multiple dialects, as well as across and within speakers. This is at least partly because previous work has either considered how speech rate is modulated by social factors and utterance length over a handful of dialects \cite{quene2008,kendall2013,bishop2018} or how speech rate varies over numerous dialects, but without controlling for utterance length \cite{coats2020}.

By looking across different varieties of the same language in the context of utterance length, this study utilises a large multi-corpus dataset of spontaneous speech to explore the structure of variation in English articulation rate (speech rate excluding pauses) -- particularly focusing on both the size and robustness of both utterance length effects and speaker factors (age, gender) across corpora -- as a window into the extent to which variability in speech timing is determined by low-level production constraints versus speaker-specific social attributes. In particular, variability in the size and direction of effects across corpora may indicate socially-conditioned variation, while relative stability in these effects may suggest that they are driven by articulatory or perceptual factors shared across varieties. This study considers the following research questions: RQ1. \emph{To what extent do the size of utterance length effects differ across corpora and speakers?}, RQ2. \emph{To what extent do corpora differ in the size of social effects (age and gender)?}, and RQ3. \emph{How do the size of social effects compare with utterance length effects?}


\section{Methods}

The data from this study comes from 116,020 utterances -- delimited by 150ms pauses -- extracted from 13 corpora of spontaneous English speech from the United Kingdom, United States, and Canada (recorded 1970-2013; 1092 speakers: 510 female), via the Integrated Speech Corpus ANalysis (ISCAN) toolkit \cite{iscan}. The corpora used in this study broadly reflect distinct regional varieties of English (most have either no reported information about speaker ethnicity, or largely are from white US, Canadian, and/or British English varieties). Speaker gender here is recorded as `male'/`female', following corpus labelling, which did not report other gender identities. Whilst each corpus differs in the particular speech style, time period, and context in which it was recorded, making it difficult to distinguish between dialectal and stylistic effects. Because of this, this study treats corpora as individual instances of speech in its own context, which may separately reflect regional and stylistic effects to their own extent.

Syllabic information was extracted using the UNISYN dialectal pronunciation lexicon \cite{unisyn}, and the measure of articulation rate was calculated as syllables per second within each force-aligned utterance \cite{kendall2013}, with a utterances defined as speech separated by at least 150ms of silence. Utterance length was calculated as the number of syllables within the utterance \cite{quene2008,bishop2018}. Utterances shorter than 3 syllables were excluded to avoid stylistic differences in very short utterances \cite{quene2008}. Utterances with articulation rates beyond $\pm$2SD from each speaker's mean articulation rate were excluded, as extreme aritculation rate values were the result of errors in forced alignment. The structure of the dataset is summarised in Table \ref{tab:data}, and details about the corpora used in the study is available at \cite{spade-osf}.

\begin{table*}[t]
  \caption{Summary of data used in final analysis: region, speech style/context, number of speakers (female), mean age (standard deviation), and number of utterances ($N$) by corpus.}
  \label{tab:data}
  \centering
  \begin{tabular}{r l l l l l}
    \toprule
    \textbf{Corpus} & \textbf{Region} & \textbf{Style} & \textbf{Speakers (F)} & \textbf{Age (SD)} & $\bm{N}$ \\
    \midrule
    1Speaker2Dialects \cite{1S2D} & NE Scotland (UK) & Sociolinguistic interviews      & 31(14)  &42 (27) & 12640 \\
    Buckeye \cite{buckeye} & Ohio (US)& Sociolinguistic interviews &                 40(20) &48 (17) &   7896 \\
    Canadian Prairies \cite{canadianprairies} & Alberta \& Manitoba (Canada) & Sociolinguistic interviews &     108(58) &41 (20) & 18379 \\
    DECTE \cite{decte} & Newcastle (UK) & Interviews, conversation &                   82(46) &37 (22) &  2928 \\
    East London & London (UK) & Sociolinguistic interviews &              57(24) &35 (27) &   327 \\
    Hastings \cite{hastings} & SE England (UK) & Sociolinguistic interviews & 31(11) &49 (25) &  6618 \\
    LUCID \cite{lucid} & London (UK) & Structured conversation task &                  40(20) &23 (0) &  2919 \\
    NEngs-Derby \cite{nengs} & Derby (UK) & Sociolinguistic interviews & 14(7) &22 (2) &  1819 \\
    NEngs-Manchester \cite{nengs} & Manchester (UK) & Sociolinguistic interviews &       27(14) &44 (24) &  2338 \\
    Raleigh \cite{raleigh} & North Carolina (US)& Spontaneous conversation &               100(50) &51 (19) & 18203 \\
    SOTC \cite{sotc} & Glasgow (UK) & Sociolinguistic interviews &                  162(63) &44 (21) & 18536 \\
    Switchboard \cite{switchboard} & Multiple locations (US) & Telephone conversations &           339(152) &36 (11) & 18855 \\
    West Virginia \cite{westvirginia} & West Virginia (US) & Sociolinguistic interviews &           61(31) &39 (22) &  4562 \\
    \midrule
    \textbf{Total} & & & 1092(510) & & 116020 \\

    \bottomrule

  \end{tabular}
  
\end{table*}

\begin{figure*}[th]
  \centering
  \includegraphics[width=0.56\linewidth]{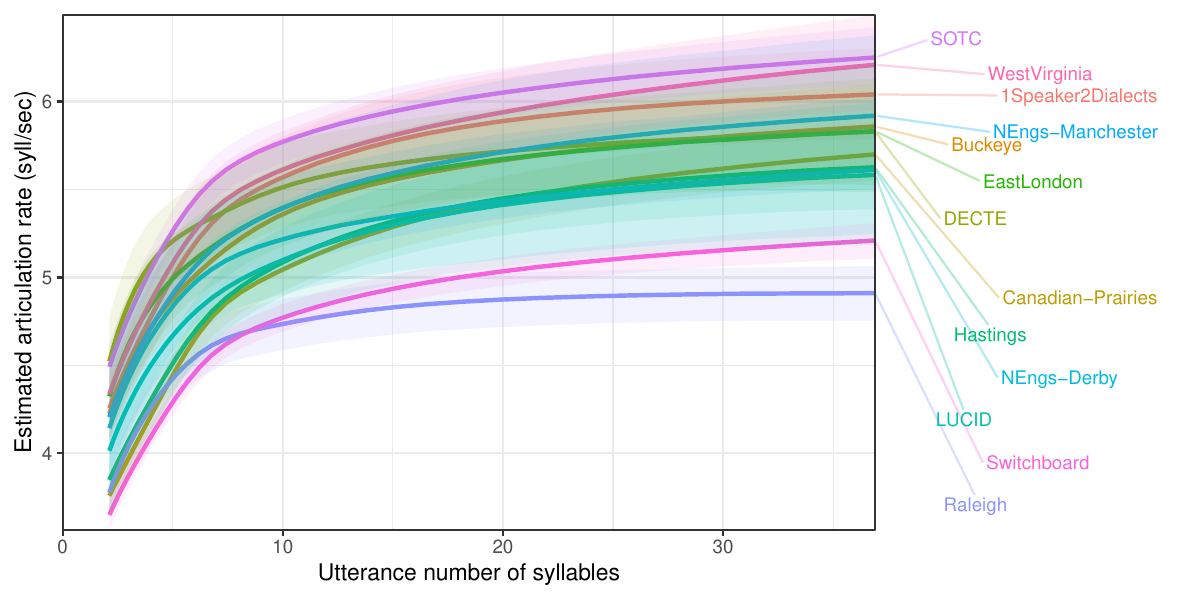}
  \includegraphics[width=0.425\linewidth]{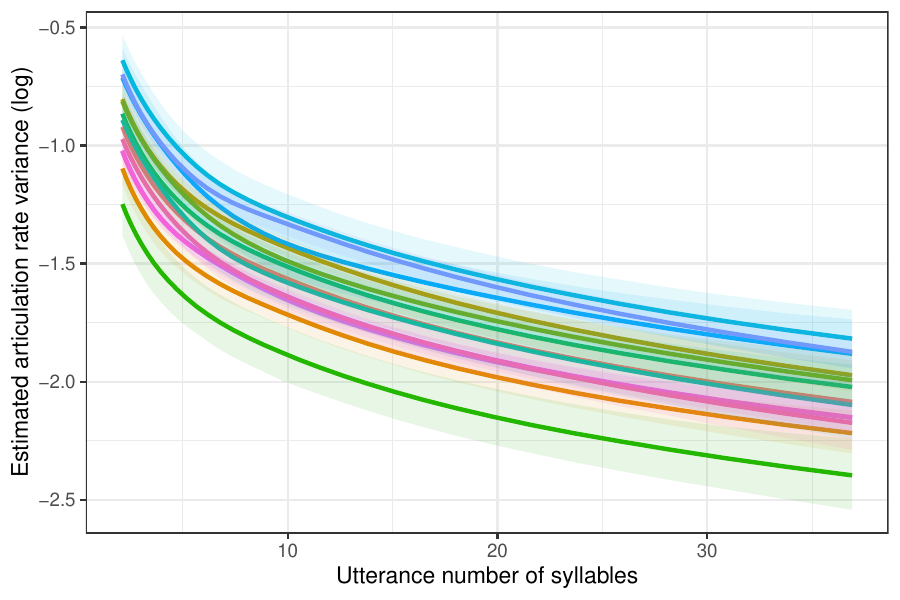}
  \caption{Estimated articulation rate (left) and variance in articulation rate (right) for each corpus as a function of utterance length. Lines indicate posterior medians with shaded areas representing 95\% Bayesian credible intervals.}
  \label{fig:utterance_length}
\end{figure*}


A distributional Bayesian multilevel model was fit to both the mean ($\mu$) and variance ($\sigma^2$) of log-transformed articulation rate using the \emph{brms} \cite{brms} interface to Stan \cite{carpenter2017stan} in \emph{R} (v4.3.1, \cite{R2022}). The log-transformed utterance length was separated into two measures: each speaker's \textbf{mean} utterance length, and the utterance-level \textbf{deviation} from that mean value, reflecting the conceptual distinction between speakers who produce shorter or longer utterances than others (on average), and for how `long' an utterance is relative to that speaker's average. Both utterance length measures were modelled as cubic splines with 5 knots to approximate the non-linear effect of utterance length on articulation rate \cite{quene2008,kendall2013}. Speaker \textbf{gender} and (z-scored) \textbf{age} were included in the model as linear predictors. By-corpus variability in the effects of utterance length mean/deviation, gender, and age, and the by-speaker variability in the utterance length deviation effect were modelled as random effects for both $\mu$ and $\sigma^2$. The model was fit with weakly-informative priors \cite{schad2021principled}: $Normal(0,2)$ for the $\mu$ intercept \& $\beta$ terms, $Exponential(1)$ for the $\mu$ group-level terms, $LKJ(1.5)$ for correlation terms, $Normal(0, log(2))$ for the $\sigma^2$ intercept \& $\beta$ terms, and $Exponential(1.4)$ for the $\sigma^2$-level group terms. The posterior distribution of the model was sampled with 2000 iterations (1000 warmup) across 4 Hamiltonian Monte Carlo chains using the \emph{cmdstanr} backend \cite{cmdstanr}, and validated with recommended diagnostic checks (e.g., posterior predictive distribution, $\hat{R}$) \cite{schad2021principled}. Code for this study is available at \cite{sr-osf}.

\section{Results}

Results are reported as summaries of the posterior distribution of the model, in terms of the posterior median, 95\% credible interval (CrI), and the posterior probability ($Pr$) of a given hypothesis (e.g., whether a particular effect's size is greater than another). Estimates of effect sizes at various levels (e.g. across speaker ages) were calculated using the \emph{emmeans} package \cite{emmeans2023}, which marginalizes across other variables. Expected posterior predictions for visualisations were extracted and summarised using the \emph{tidybayes} package \cite{tidybayes2023}.

\begin{figure*}[th]
  \centering
  \includegraphics[width=\linewidth]{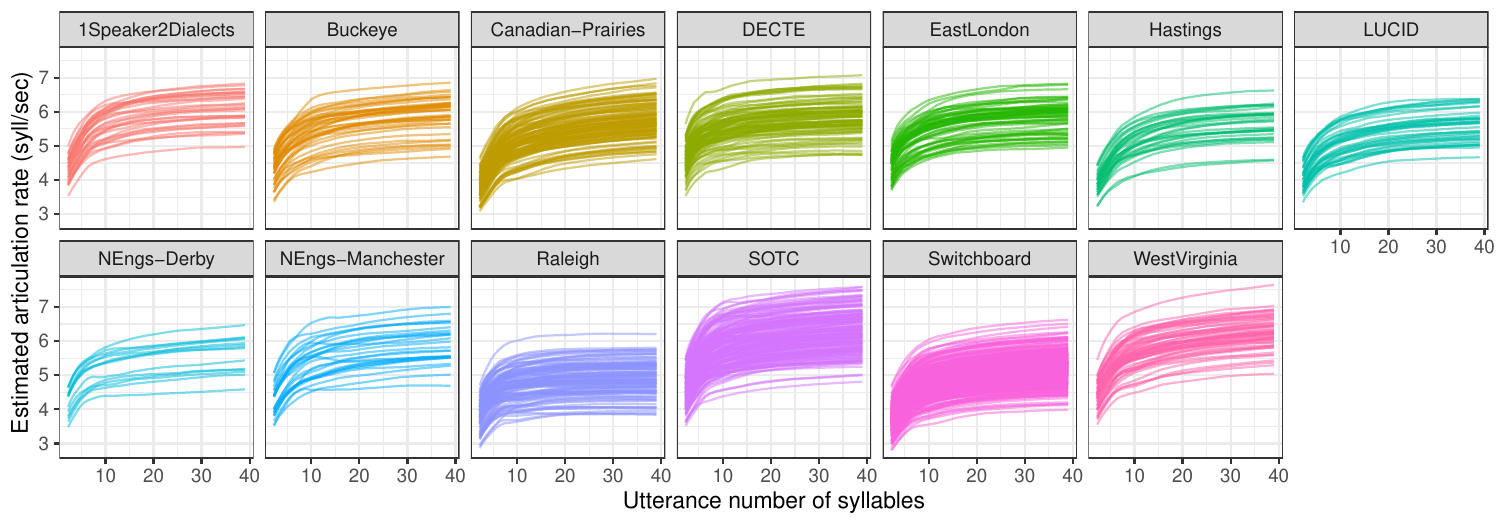}
  \caption{Estimated articulation rate for each speaker (faceted by corpus) as a function of utterance length (one line per speaker). Values indicate posterior medians.}
  \label{fig:speaker}
\end{figure*}

With respect to RQ1, the length of the utterance has a significant effect on both the mean and variance of articulation rate, though the effect of utterance length does not appear to meaningfully differ between corpora (Fig.~\ref{fig:utterance_length}). Corpora ($\hat{\sigma}_{corpus}$ = $0.07$, CrI = [$0.04$,$0.13$]; Fig.~\ref{fig:utterance_length} left) and speakers ($\hat{\sigma}_{speaker}$ = $0.09$, CrI = [$0.09,0.10$]; Fig.~\ref{fig:speaker}) differ substantially in their average articulation rate, but there is little evidence that variation across individual speakers is necessarily greater than the variation between corpora ($\hat{\sigma}_{speaker} > \hat{\sigma}_{corpus}$ = $0.02$, CrI = [$-0.02,0.05$], $Pr$ = $0.87$), and speakers (within-dialect) themselves vary little in how utterance length modulates articulation rate (Fig.~\ref{fig:speaker}). Similar to changes in average articulation rate, corpora differ in their overall \emph{variance} in articulation rate ($\hat{\sigma}_{\sigma corpus}$ = $0.19$, CrI = [$0.12$,$0.30$]), though appear not to differ in how articulation rate variance reduces as a function of utterance length (Fig.~\ref{fig:utterance_length} right).

\begin{figure}[h]
  \centering
  \includegraphics[width=\linewidth]{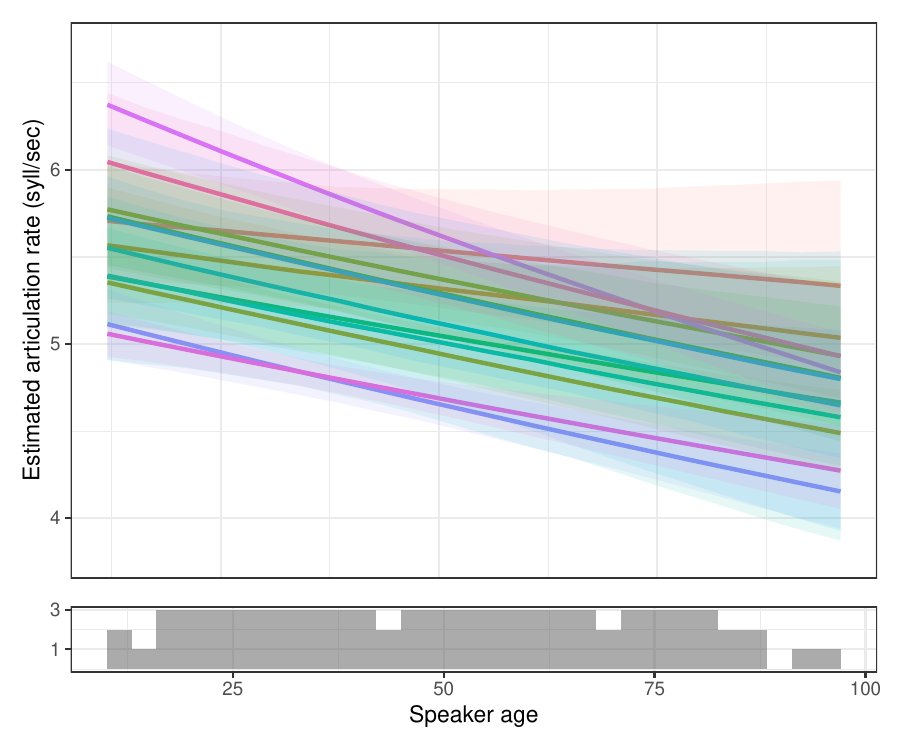}
  \caption{Top: estimated articulation rate as a function of speaker age across corpora, held at mean utterance length. Values indicate posterior medians (one line per dialect) with 95\% Bayesian credible intervals. Bottom: histogram of speaker ages.}
  \label{fig:age}
\end{figure}

\begin{figure}[h]
  \includegraphics[width=\linewidth]{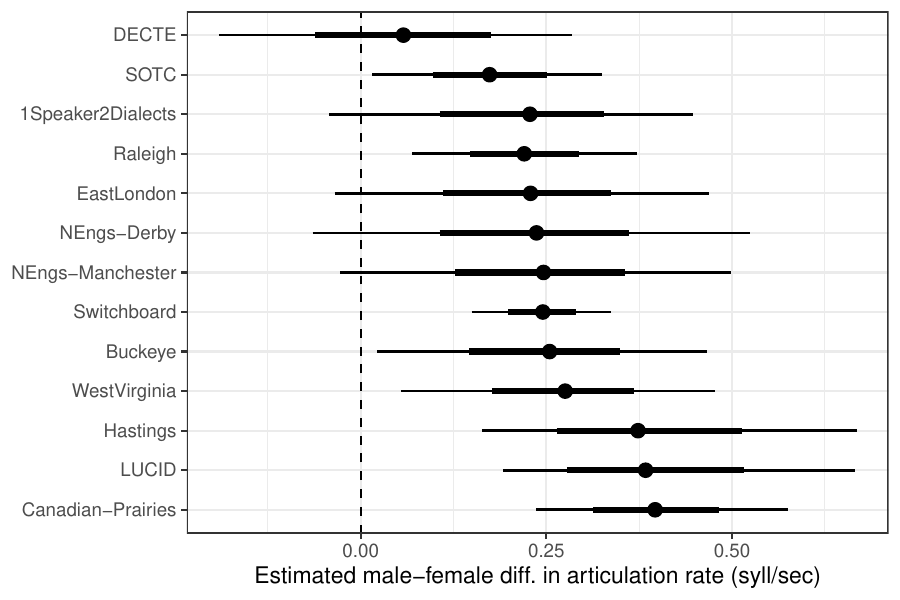}
  \caption{Estimated male-female difference in articulation rate for each corpus, held at mean utterance length. Points and lines indicate posterior medians with 95\% Bayesian credible intervals, with dashed vertical line indicating null gender difference.}
  \label{fig:gender}
\end{figure}

Looking at the effects of social factors (RQ2), age is found to have a negative effect on articulation rate, meaning that older speakers are predicted to speak more slowly than younger speakers ($\hat{\beta}_{age}$ = $-0.04$, CrI = [$-0.05,-0.02$]). This negative effect itself differs little across corpora ($\hat{\sigma}_{age}$ = $0.02$, CrI = [$0.01$,$0.04$], with articulation rate decreasing by approximately 1 syllable per second across the age range (Fig.~\ref{fig:age}). Articulation rate also differs by speaker gender, where female speakers are predicted to speak approximately 0.25 syllables per second more slowly than male speakers ($\hat{\beta}_{gender}$ = $0.05$, CrI = [$0.03,0.07$]; Figure \ref{fig:gender}). Figure \ref{fig:gender} also illustrates variation in the size of the gender effect across corpora ($\hat{\sigma}_{gender}$ = $0.03$, CrI = [$0.01$,$0.06$]), where the majority of corpora with estimated effect sizes overlapping with 0 (DECTE, 1Speaker2Dialects, East London, NEngs-Derby/Manchester) are also those with the fewest number of utterances or speakers (Tab.~\ref{tab:data}).


The size of social factor effects on articulation rate are relatively small compared with the effect of utterance length (RQ3); while gender and age affect articulation rate by 0.21-0.24 syll/sec (Fig.~\ref{fig:gender}) and  0.7-0.8 syll/sec (Fig.~\ref{fig:age}) respectively, articulation rate increases by 2.1-2.4 syll/sec between the shortest and longest utterances (Fig.~\ref{fig:utterance_length} left). These findings demonstrate that the social effects of age and gender are still present once utterance length has been accounted for, but the size of these effects are much smaller in magnitude than that of utterance length.

\section{Discussion}


Speech rate has been shown to vary according to both socially-structured categories such as dialect, age, and gender \cite{jacewicz2010rate,fougeron2021}, as well as being constrained by speech production processes, where longer utterances result in both increased speech rate as well as less variation in rate over longer utterances \cite{quene2008,kendall2013,bishop2018}. The goal of this study is to explore at a larger scale, how variability in speech timing -- specifically speech rate -- is structured along both sets of factors, namely how both social factors and utterance length interact in conditioning speech rate variation, and in the relative prominence of social effects once utterance length has been accounted for. These questions are addressed through the analysis of a large dataset of spontaneous English speech, created through the combination of data from multiple speech corpora, for which information about social factors (dialect, gender, age) and utterance length, is accessible. 


Utterance length was found to be by far the largest predictor of articulation rate variation, where longer utterances were produced at faster speech rates well as less variance in rate, following observations in previous work, albeit for a few dialects of Dutch and US English respectively \cite{quene2008,kendall2013}. While speech rate varies to an extent across corpora and individual speakers, the trajectory of the utterance length effect itself, however, did not meaningfully differ across these groups, indicating that individuals modulate their articulation rates based on utterance length in largely similar ways. The relative invariance in utterance length effects indicates that these may be driven by constraints on articulation: articulatory gestures can only be performed so quickly, even after accounting for gestural undershoot \cite{byrd1996}. Perceptual constraints also likely play a role in how articulation rate is modulated in this way: speakers are still required to produce intelligible speech and allow for temporally-conditioned phonetic cues to remain perceptible \cite{Lindblom1990,smiljanic2008,kawahara2022}.


While the effects of social factors were present and moved in their expected directions (such as older speakers and female speakers producing slower speech rates), the influence of social effects was much smaller in size than that of the utterance length effects. In this sense, the role of social factors on conditioning articulation rate variation is meaningful, but their effects must be placed in context of the greater influence of utterance length, again confirming previous observations, but now from a much larger base of dialect data \cite{quene2008,jacewicz2010rate,kendall2013}. Given the relatively small effects of social factors, it remains an open question about the extent to which socially-structured differences in speech rate are themselves socially perceptible. The social indexing of speech is inherently multi-dimensional -- for example, speaker dialect may be cued by spectral information and dialect-specific durational contrasts \cite{fridland2014} -- and so how the effects of social factors relate to broader social stereotypes and perceptions about speech rate remains a topic of further investigation.


\section{Acknowledgements}
The authors wish to thank the SPADE Data Guardians, Rachel Macdonald, Michael McAuliffe, Vanna Willerton, and audience of the 2024 Colloquium of the British Association of Academic Phoneticians. This research was supported by a T-AP Digging into Data award, in the form of the following grants: ESRC Grant \#ES/R003963/1, NSERC/CRSNG Grants \#RGPDD-501771-16 and \#RGPIN-2023-04873, SSHRC/CRSH Grant \#869-2016-0006, and NSF Grant \#SMA-1730479, Canada Research Chair \#CRC-2023-00009 (MS); and a British Academy Postdoctoral Fellowship awarded to JT.

\bibliographystyle{IEEEtran}
\bibliography{JTanner}

\end{document}